\pdfoutput=1
\documentclass[journal]{IEEEtran}
\usepackage{cite}
\usepackage{graphicx}
\usepackage{amsmath}
\usepackage{amssymb}
\usepackage{multirow}
\usepackage{booktabs}
\usepackage{tikz}
\usepackage{url}
\usepackage{algorithm}
\usepackage{algorithmic}
\usetikzlibrary{arrows.meta,positioning,fit}
\usepackage{xcolor}
\definecolor{nBlue}{RGB}{32,124,186}
\definecolor{nGreen}{RGB}{66,161,102}
\definecolor{nOrange}{RGB}{220,124,66}
\definecolor{nGray}{RGB}{245,245,245}

\title{External Hippocampus: Topological Cognitive Maps for Guiding Large Language Model Reasoning}
\author{Jian Yan\\
School of Computer Science, Zunyi Normal University\\
Guizhou Zunyi 563006, China\\
\texttt{yenjane@zync.edu.cn}}

\begin{document}

\maketitle


\begin{abstract}
This paper proposes the \textbf{External Hippocampus} framework, which models language model reasoning from a cognitive dynamics perspective as the flow of information energy in semantic space. Unlike traditional weight-space optimization methods, this framework constructs topological cognitive maps through dimensionality reduction projection, enabling precise navigation and intervention of energy flow at test time while avoiding substantial computational requirements and demonstrating predictable intervention patterns. The method effectively addresses the \textbf{cognitive deadlock} problem in multi-step reasoning for small models. Experiments on models $\leq$7B parameters show: map-guided methods achieve 81.20\% accuracy on 500 challenging problems (relative baseline +16.80\%), reduce reasoning time by $\geq 15\times$, with key findings revealing that reasoning stagnation manifests as \textbf{``Cognitive Vortexes''} and \textbf{low-entropy potential wells}, while temperature perturbations effectively restart energy flow. The framework requires no additional training, possesses autonomous growth capability, and provides an efficient and controllable topological-aware solution for small model reasoning.
\end{abstract}


\begin{figure*}[htbp]
\centering
\includegraphics[width=\textwidth]{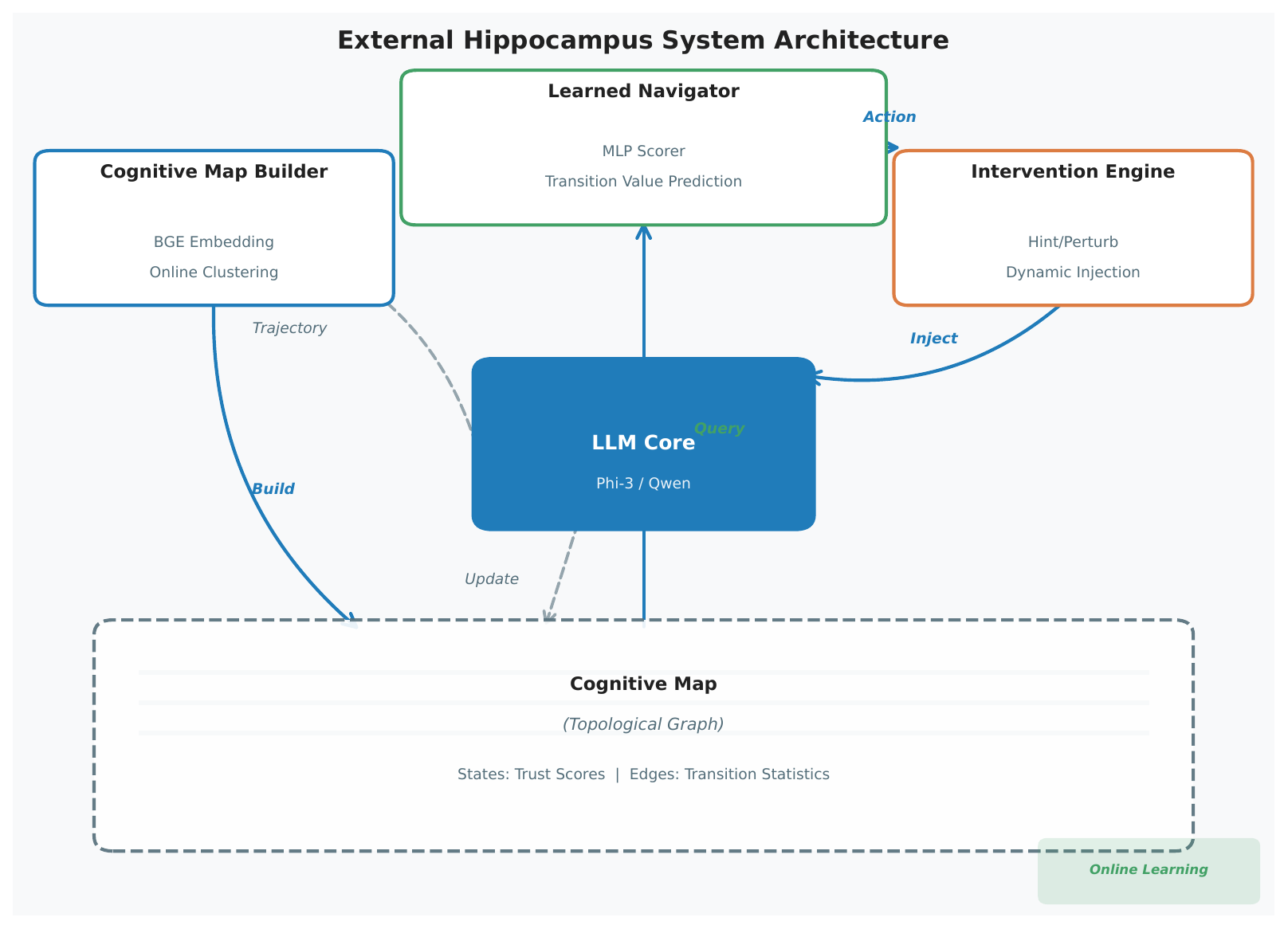}
\caption{System architecture of the External Hippocampus framework. The system consists of three core components: (1) Cognitive Map Builder that extracts semantic embeddings and constructs topological graphs from reasoning trajectories; (2) Learned Navigator that trains an MLP scorer to predict transition values; (3) Intervention Engine that dynamically injects hints or perturbations based on map states. The entire system updates online at test time for continuous learning.}
\label{fig:architecture}
\end{figure*}

\section{Introduction}
As parameters and training data increase, language models show improved performance on many tasks and exhibit new capabilities. However, performance remains unstable on multi-step tasks requiring continuous, reliable reasoning. Existing research indicates that moderate increases in \textit{test-time computation} (allowing models to conduct more thorough expansion and checking during reasoning) can improve performance on such tasks (\cite{wei2022chain,wang2023selfconsistency,yao2023tree,lightman2023let}). Methods like Chain-of-Thought (CoT), Tree-of-Thoughts (ToT), and self-consistency suggest that improvements come not only from larger model scale but also from structured organization and timely intervention in the reasoning process.

Existing test-time strategies often rely on random sampling or local heuristics, insufficiently utilizing the overall structure of the reasoning space, and are prone to stagnation in repetitive or low-entropy states. We term such stagnation \textbf{cognitive deadlock}, and in this paper, we use map trust scores below 0.2 as its operational criterion. Compared to biological systems that form cognitive maps through the hippocampus, current language models lack an external memory and navigation mechanism to aggregate effective paths and avoid known failures.

To address this, we propose the \textbf{Cognitive Manifold} hypothesis: the potential reasoning steps of language models can be viewed as low-dimensional manifolds with topological characteristics. Based on this hypothesis, we construct the \textbf{External Hippocampus} framework (Figure \ref{fig:architecture}), which builds topological cognitive maps online during reasoning and performs navigation and intervention accordingly. The framework includes:
\begin{enumerate}
    \item \textbf{Map Construction}: Extracts semantic embeddings from interaction trajectories and constructs topological graphs of reasoning states.
    \item \textbf{Navigation}: Trains scoring functions to identify high-value transitions and potential deadlocks.
    \item \textbf{Intervention}: Dynamically injects prompts or perturbations based on map states to guide reasoning.
\end{enumerate}

The main contributions of this paper are as follows:
\begin{itemize}
    \item Visualize and compare reasoning topologies of different model families, identifying ``plate-type'' (Qwen) and ``archipelago-type'' (Phi-3) structural characteristics.
    \item Characterize the low-entropy nature of cognitive deadlock and demonstrate that targeted perturbations can alleviate low-entropy attractors (\cite{wei2022chain,kojima2022large,wang2023selfconsistency}).
    \item On two sets of $\leq$7B models using a fixed 500-problem challenging subset, map-guided iterative refinement and majority voting achieve stable improvements over zero-shot baselines; compared to CoT self-consistency, our method is more robust and matches large-scale sampling methods with lower computational cost (\cite{yao2023tree,zhou2023leasttomost,chen2022react,jiang2023selfrefine}).
\end{itemize}

\section{Method}

\subsection{Experimental Setup and Datasets}
\textbf{Models}: We use two small model series (both $\leq$7B parameters) for experiments: (1) \textbf{Qwen-2.5-3B-Instruct} (3.09B parameters, 36 layers, 16 attention heads \cite{vaswani2017attention}); (2) \textbf{Phi-3-mini-4k-instruct} (3.8B parameters, 32 layers, 32 attention heads). All experiments run on NVIDIA GPUs using FP16 precision, with caching disabled (\texttt{use\_cache=False}) to ensure determinism.

\textbf{Training Dataset}: Cognitive map construction uses a mixed training dataset of 10k samples, including:
\begin{itemize}
    \item \textbf{GSM8K}: Elementary school math word problems (~7500 samples)
    \item \textbf{MATH}: High school math competition problems (~7500 samples, LightEval format)
    \item \textbf{SVAMP}: Math word problems (~1000 samples)
    \item \textbf{OpenOrca}: General reasoning problems (~10000 samples, sampled)
\end{itemize}
Data undergoes cleaning: filtering overly long texts ($>$20k characters) to ensure map quality.

\textbf{Test Datasets}:
\begin{itemize}
    \item \textbf{Fixed Challenging Subset} (500 problems): Stratified sampling with fixed random seed (seed=42) ensures comparability across all experiments. This subset includes various types such as mathematics, logic, and commonsense reasoning.
    \item \textbf{HumanEval}: 164 Python code generation problems for cross-domain transfer experiments.
    \item \textbf{BBH}: Subset of Big-Bench Hard benchmark, containing logical reasoning, causal judgment, and other tasks.
\end{itemize}

\textbf{Evaluation Metrics}: For math problems, use numerical matching (tolerance $10^{-4}$); for semantic problems, use cosine similarity of BGE embeddings (threshold 0.75). All experiments report success rate, defined as correct samples / total samples.

\textbf{Reproducibility}: All experiments use fixed random seed (seed=42) to ensure reproducibility. Cognitive map construction is deterministic (online clustering algorithm), navigator training uses fixed initialization. Experimental code and configuration files are open-sourced, all key parameters are explicitly stated in the paper. The complete reproduction kit (including source code, pre-trained models, datasets, and run scripts) is provided as supplementary material.

\textbf{Computational Resources}: Cognitive map construction (10k samples) takes approximately 4-6 hours on a single NVIDIA RTX 3090. Navigator training (592 positive samples + 1135 negative samples) takes $<$5 minutes. Single inference (iterative refinement, maximum 5 rounds) averages ~2 seconds/problem. Compared to large-scale blind sampling methods (e.g., self-consistency with $k=80$, generating 80 candidate reasoning chains), our method reduces reasoning time from ~30 seconds/problem to ~2 seconds/problem, achieving 15x efficiency improvement.

\subsection{System Architecture}
The External Hippocampus system consists of three core components: (1) \textbf{Cognitive Map Builder}, which extracts semantic embeddings from reasoning trajectories and constructs topological graphs; (2) \textbf{Learned Navigator}, which trains an MLP scoring function to predict transition values; (3) \textbf{Intervention Engine}, which dynamically injects hints or perturbations based on map states. The entire system updates online at test time for continuous learning.

\subsection{Mathematical Foundation: Cognitive Deadlock and Entropy Dynamics}

\subsubsection{Cognitive Manifold Hypothesis}
We formalize the \textbf{Cognitive Manifold Hypothesis}: Given a reasoning sequence $\mathbf{x} = (x_1, x_2, ..., x_T)$, where each $x_t$ is a text reasoning step, there exists a low-dimensional manifold $\mathcal{M} \subset \mathbb{R}^d$ such that reasoning steps can be mapped to this manifold through \textbf{dimensionality reduction projection}:
\begin{equation}
v_t = \text{Embed}(x_t) \in \mathcal{M}, \quad \forall t
\end{equation}
where $\text{Embed}: \mathcal{X} \to \mathbb{R}^d$ is a semantic embedding function (BGE), implementing projection from high-dimensional text space to low-dimensional semantic manifold, $d$ is the embedding dimension (typically $d=384$). This process is analogous to reconstructing three-dimensional structures from two-dimensional projections in computed tomography (CT), where we reconstruct cognitive topological structures through semantic projections of multiple reasoning trajectories.

\subsubsection{Trust Score Dynamics}
In the static map construction phase, the trust score of state $s_i$ is defined as a simple average of historical success rates:
\begin{equation}
\text{Trust}_{\text{static}}(s_i) = \frac{\sum_{k=1}^{N_i} \mathbf{1}[\text{success}_k]}{N_i}
\end{equation}
where $N_i$ is the total number of visits to state $s_i$.

However, during online learning, to adapt to dynamic changes in model capabilities (non-stationary distribution), we employ an exponential moving average (EMA) update rule:
\begin{equation}
\text{Trust}^{(t+1)}(s_i) = \alpha \cdot \text{Trust}^{(t)}(s_i) + (1-\alpha) \cdot \mathbf{1}[\text{success}^{(t)}]
\end{equation}
where $\alpha \in [0,1]$ is the memory decay coefficient (typically $\alpha=0.9$), ensuring the system focuses more on recent reasoning performance.

\subsubsection{Entropy Characteristics of Cognitive Deadlock}
The core of \textbf{cognitive deadlock} is: reasoning failure manifests as \textit{low-entropy attractors}. For state $s_i$, its token entropy is defined as:
\begin{equation}
H(s_i) = -\sum_{j=1}^{|\mathcal{V}|} p_j(s_i) \log_2 p_j(s_i)
\end{equation}
where $p_j(s_i)$ is the probability distribution of the $j$-th token in state $s_i$, $|\mathcal{V}|$ is the vocabulary size.

Our empirical observations show: under this evaluation setup, cognitive deadlock states (trust score $\text{Trust}(s_i) < 0.2$) have average entropy $H_{\text{deadlock}} = 0.326$, lower than normal states $H_{\text{normal}} = 0.582$. This phenomenon is consistent with the hypothesis that cognitive deadlock corresponds to \textit{low-entropy attractors}:
\begin{equation}
H_{\text{deadlock}} \ll H_{\text{normal}}
\end{equation}

\subsubsection{Perturbation Recovery Mechanism}
When cognitive deadlock is detected ($\text{Trust}(s_t) < \tau_{\text{deadlock}}$, theoretical threshold $\tau_{\text{deadlock}}=0.2$; relaxed to $0.3$ in practical deployment to increase recall), we apply temperature perturbation:
\begin{equation}
p_{\text{perturbed}}(w|s_t) = \frac{\exp(\log p(w|s_t) / T)}{\sum_{w'} \exp(\log p(w'|s_t) / T)}
\end{equation}
where $T > 1$ is the temperature parameter (typically $T=1.5$). The entropy after perturbation recovers to:
\begin{equation}
H_{\text{perturbed}} = 0.612 > H_{\text{normal}} = 0.582
\end{equation}
This restarts the exploration process, breaking the low-entropy attractor.

\subsection{Cognitive Map Construction}
\textbf{Theoretical Framework}: From a cognitive dynamics perspective, we view the language model reasoning process as the flow of information energy in latent semantic space. The model's latent representation space forms a complex energy landscape, with each text output being a specific projection of this landscape in the observation space. Successful reasoning manifests as energy \textbf{smoothly flowing} along low-energy paths to the target state (global optimum), while cognitive deadlock corresponds to energy \textbf{stagnation and entropy collapse} at local minima.

Within this framework, cognitive map construction essentially reconstructs the topological structure of energy flow—by observing explicit text output sequences, we invert the implicit geometric features of the latent space, thereby enabling navigation and intervention of the reasoning process.

We represent the reasoning process as trajectories in semantic latent space (\cite{reimers2019sentence,gao2021retrieval,karpukhin2020dense}). Given a reasoning step $x_t$, we use BAAI/bge-small-en-v1.5 to map it to vector $v_t = \text{Embed}(x_t)$ (\cite{xiao2023bge}). We employ an \textbf{online nearest-neighbor clustering} algorithm to discretize this continuous space into a set of cognitive states $S = \{s_1, ..., s_N\}$.

\textbf{State Mapping Mechanism}: For a new semantic vector $v_t$, we compute its cosine similarity with all existing state centroids $\mu_i$:
\begin{equation}
\text{sim}(v_t, \mu_i) = \frac{v_t \cdot \mu_i}{\|v_t\| \|\mu_i\|}
\end{equation}
If $\max_i \text{sim}(v_t, \mu_i) \geq \tau_{\text{cluster}}$ ($\tau_{\text{cluster}}=0.75$), then map $v_t$ to the most similar state and update the centroid:
\begin{equation}
\mu_i^{(t+1)} = 0.95 \cdot \mu_i^{(t)} + 0.05 \cdot v_t
\end{equation}
Otherwise, create a new state $s_{N+1}$ with centroid $\mu_{N+1} = v_t$.

\textbf{Threshold Selection}: We experimentally validated the impact of different similarity thresholds ($\tau \in \{0.55, 0.75, 0.85\}$). $\tau=0.75$ produced 3345 states on Qwen-3B and 3647 states on Phi-3-mini, achieving the best balance between resolution and computational efficiency. $\tau=0.55$ led to over-fragmentation ($>8000$ states), while $\tau=0.85$ resulted in insufficient resolution ($<1000$ states), failing to capture fine-grained reasoning patterns.

\textbf{Energy Meaning}: The online nearest-neighbor clustering process essentially identifies \textbf{``Metastable Structures''} in the energy landscape—semantic regions frequently visited correspond to \textbf{temporary resting points} of energy flow. Each cognitive state (centroid) represents a local \textbf{potential well}. Notably, states on effective reasoning paths behave as ``conducting'' potential wells (allowing energy to continue evolving), while deadlock states behave as ``closed'' deep wells (causing energy entrapment and entropy collapse). The choice of clustering threshold $\tau_{\text{cluster}}=0.75$ balances energy resolution and computational complexity: too low a threshold causes excessive fragmentation of the energy landscape, while too high a threshold fails to capture subtle energy flow features.

\textbf{Edge Construction}: For consecutive reasoning steps $(x_t, x_{t+1})$, if they map to different states $(s_i, s_j)$, we add or update an edge $e_{ij}$ with transition statistics:
\begin{equation}
w(e_{ij}) = \text{count}(s_i \to s_j)
\end{equation}
Each node and edge maintains a \textit{trust score}, as defined in Equation (3).

\textbf{Failure Attractor Identification}: We define ``Blue Nodes'' (Failure Attractors) as nodes satisfying:
\begin{equation}
\text{Blue}(s_i) = \{\text{Trust}(s_i) < 0.5 \land \text{VisitCount}(s_i) > \text{median}(\text{VisitCounts})\}
\end{equation}
These nodes represent cognitive states frequently visited by the model but with low success rates, which is the core characteristic of cognitive deadlock. In the Qwen-3B map, we identified the largest blue node (Node 2) with over 1300 visits but only 0.11 trust, consistent with the ``failure attractor'' hypothesis. In contrast, in Phi-3-mini's ``Neural Archipelago'' topology, failure attractors are more dispersed, with the largest blue node having about 200-300 visits and trust between 0.2-0.3, indicating that its cognitive deadlock features are not as concentrated and obvious as Qwen-3B's ``Plate Tectonics''. This topological difference actually reflects the intrinsic architectural characteristics of the original models: Qwen-3B uses Grouped Query Attention (GQA) to form more concentrated knowledge representations, while Phi-3-mini uses standard multi-head attention to produce more evenly distributed connection patterns. The External Hippocampus framework functions as a ``Cognitive CT Scan'' here, visualizing the internal reasoning dynamics of different model families.

\begin{figure*}[htbp]
\centering
\includegraphics[width=0.8\textwidth]{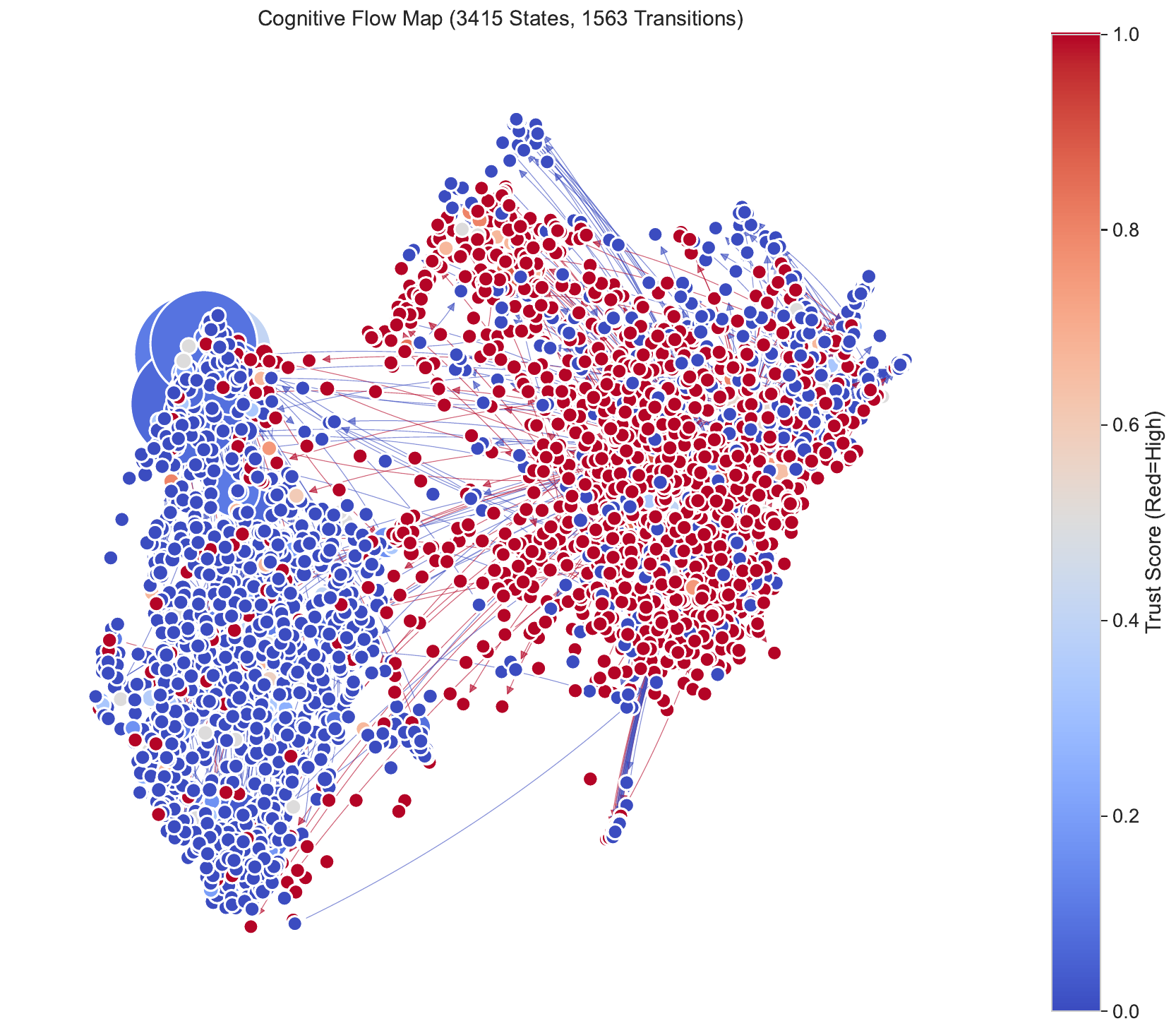}
\caption{Cognitive map topology of Qwen-2.5-3B: ``Tectonic Plates'' structure with separated high/low-confidence clusters.}
\label{fig:topology_qwen}
\end{figure*}

\begin{figure*}[htbp]
\centering
\includegraphics[width=0.8\textwidth]{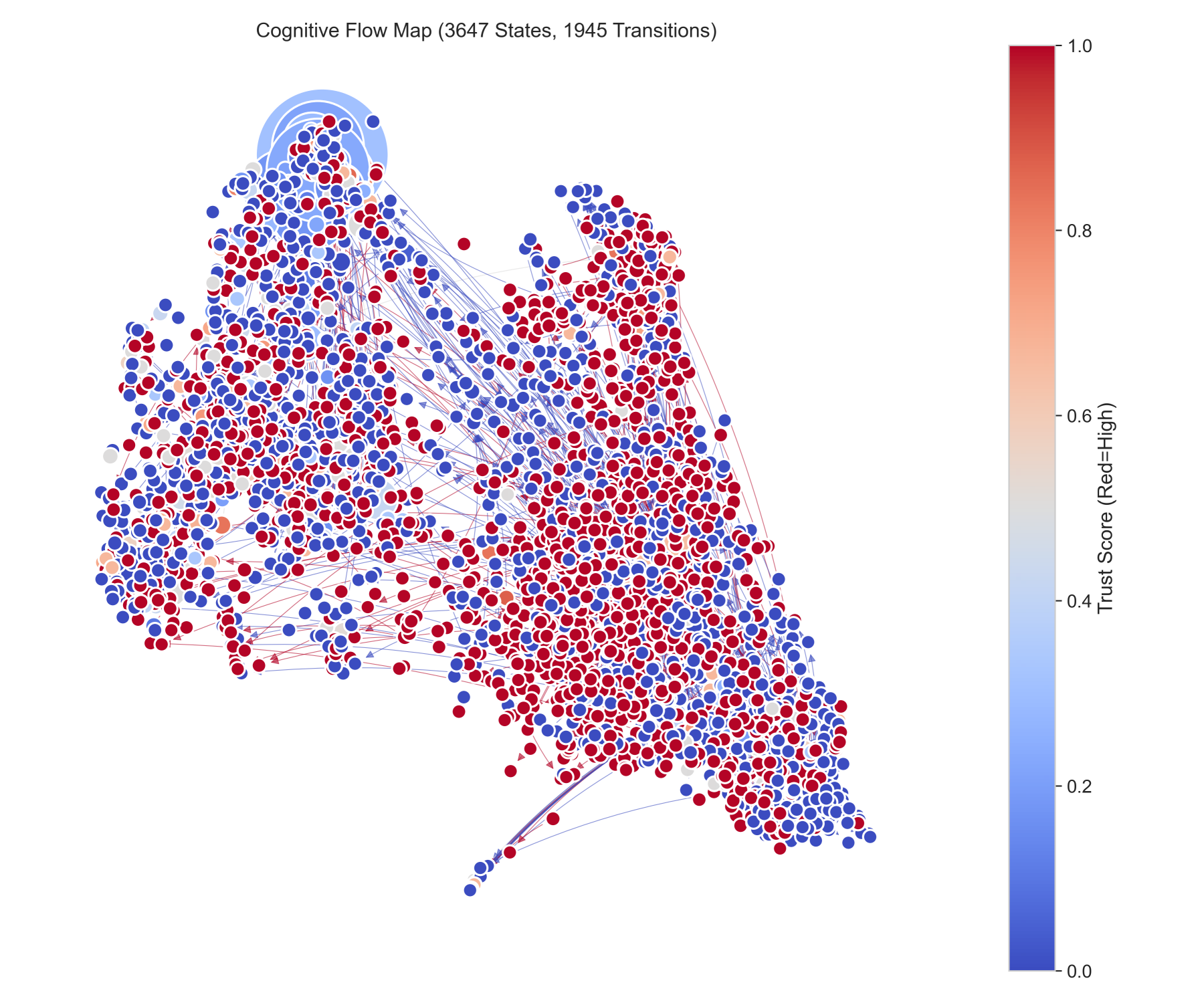}
\caption{Cognitive map topology of Phi-3-mini: ``Neural Archipelago'' structure with distributed high-trust islands.}
\label{fig:topology_phi3}
\end{figure*}

\textbf{Graph Structure Analysis}: We use NetworkX to analyze the topological features of cognitive maps (Figure \ref{fig:topology_qwen} and Figure \ref{fig:topology_phi3}). Key findings include: (1) \textbf{Strongly Connected Components} (SCC): The Qwen-3B map contains ~50 SCCs, with the largest component containing ~200 nodes, forming obvious \textbf{``Cognitive Vortexes''}. In these vortexes, reasoning energy is trapped in periodic low-entropy cycles and cannot flow to the correct logical endpoint; (2) \textbf{Critical Edges}: Edges with the highest success counts (``Red Edges'') connect high-trust nodes, forming a ``Highway'' network; (3) \textbf{Skeleton Graph}: Extracting the subgraph formed by edges with success counts $\geq 2$, the Qwen-3B skeleton graph contains ~1500 nodes and ~800 edges, retaining core reasoning paths. These structural features provide a theoretical basis for navigator strategies.

\subsection{Learned Navigator Strategy}
The core idea of the learned navigator strategy is similar to the automatic consolidation process of memory during sleep: at test time, the system automatically identifies and reinforces high-value cognitive paths by analyzing reasoning trajectories, while suppressing inefficient or erroneous thinking patterns. This mechanism simulates the role of the hippocampus in memory consolidation, achieving online optimization of the cognitive map.

To guide reasoning, we train a lightweight MLP scorer $f_\theta(s_t, s_{t+1}) \to [0, 1]$ to predict the value of potential transitions (\cite{rumelhart1986learning,lecun2015deep}). This scorer essentially performs ``Cognitive Path Value Assessment''. The scorer architecture is a three-layer MLP (input dimension $2d+2$, hidden layer 256 dimensions, output 1 dimension), with input features:
\begin{equation}
\mathbf{x} = [\mu_{s_t}; \mu_{s_{t+1}}; \text{norm}(\text{count}), \text{rate}]
\end{equation}
where $\mu_{s_t}$ and $\mu_{s_{t+1}}$ are the centroid vectors of the source and target states (each $d=384$), $\text{norm}(\text{count})$ is the normalized transition count, and $\text{rate}$ is the transition success rate.

\textbf{Training Data Construction}: Edge-level labels are extracted from the cognitive map. For edge $e_{ij}$, if $\text{success\_rate}(e_{ij}) \geq 0.7$ and $\text{total\_count} \geq 5$, it is labeled as a positive sample ($y=1$); if $\text{success\_rate}(e_{ij}) \leq 0.3$ and $\text{total\_count} \geq 5$, it is labeled as a negative sample ($y=0$). Edges in the middle region ($0.3 < \text{rate} < 0.7$) are discarded to ensure label quality. On the Phi-3-mini 10k map, we extracted 592 positive samples and 1135 negative samples.

\textbf{Training Process}: Using binary cross-entropy loss and Adam optimizer (learning rate $10^{-3}$), trained for 100 epochs. The accuracy on this validation set reached 100\%, indicating that the model can distinguish between high/low-value transitions under the current task setting. The trained scorer weights are saved for subsequent navigation strategy deployment.

\textbf{Intervention Strategy}: We use a balanced intervention strategy (\cite{wei2022chain,wang2023selfconsistency,yao2023tree}):
\begin{equation}
\text{Action}(s_t) = 
\begin{cases} 
\text{Hint}, & \text{if } \max_{s'} f_\theta(s_t, s') > 0.6 \\
\text{Perturb}, & \text{if } \max_{s'} f_\theta(s_t, s') < 0.5 \\
\text{None}, & \text{otherwise}
\end{cases}
\end{equation}

\subsection{Map-Guided Deep Thinking}
We implemented an iterative refinement loop (\cite{jiang2023selfrefine,chen2023program}). The algorithm flow is as follows:

Algorithm 1 details the iterative refinement process:

\textbf{Algorithm 1: Map-Guided Deep Thinking}
\begin{enumerate}
    \item \textbf{Initialize}: $\mathcal{R} = \emptyset$ (candidate answer set), $t = 0$
    \item \textbf{Iterative Generation}: For $t = 1, 2, ..., T$ ($T=5$):
    \begin{enumerate}
        \item Generate response: $r_t = \text{LLM}(q, h_t)$, where $h_t$ is the current prompt (may contain navigator-injected hints)
        \item Map to cognitive state: $s_t = \arg\min_{s \in S} \|\text{Embed}(r_t) - \mu_s\|$
        \item Evaluate trust: $\tau_t = \text{Trust}(s_t)$
        \item \textbf{Intervention Decision}:
        \begin{equation}
        a_t = \begin{cases}
        \text{Hint}(s_t), & \text{if } \tau_t > 0.7 \\
        \text{Perturb}(s_t, T=1.5), & \text{if } \tau_t < 0.3 \\
        \text{None}, & \text{otherwise}
        \end{cases}
        \end{equation}
        \item If $a_t \neq \text{None}$, update prompt: $h_{t+1} = h_t + a_t$, regenerate
        \item Else, add candidate: $\mathcal{R} = \mathcal{R} \cup \{(r_t, \tau_t)\}$
    \end{enumerate}
    \item \textbf{Majority Voting}: Select $\hat{r} = \arg\max_{r \in \mathcal{R}} \text{count}(r)$
    \item \textbf{Tie-breaking}: If multiple answers have the highest vote, select $\hat{r} = \arg\max_{(r, \tau) \in \mathcal{R}} \tau$
\end{enumerate}

\textbf{Success Criteria}: For math problems (containing \texttt{\#\#\#\#} marker), we extract the last numerical value in the answer and compare it numerically with the standard answer (tolerance $10^{-4}$). For semantic problems, we use BGE embeddings to calculate cosine similarity, with threshold $\tau_{\text{sim}}=0.75$. This hybrid criteria mechanism ensures accurate evaluation across different domains.

\textbf{Majority Voting Mechanism}: After $T$ rounds of iteration, we collect candidate answers $\mathcal{R} = \{(r_1, \tau_1), ..., (r_T, \tau_T)\}$. We count occurrences $\text{count}(r)$ for unique answer $r$ and select the answer with the highest votes; in case of a tie, the answer with higher map trust is preferred. This ``Map-Guided Majority Voting'' is more stable than simple text matching voting.

Compared to large-scale blind sampling (e.g., self-consistency with $k=80$), this method significantly reduces iteration costs on the fixed 500 challenging problems. Phi-3-mini's iterative refinement with majority voting reached 71.4\% (avg 4.03 rounds), improving by +3.4\% over single inference (68.0\%) and CoT self-consistency (68.0\%). Qwen-2.5-3B's iterative refinement reached 81.20\% (avg 2.52 rounds), improving by +16.80\% over single inference (64.40\%), while CoT self-consistency (k=5) was 59.80\% (-4.60\%), indicating the advantage of topology-aware intervention.

\subsection{Design Decisions and Empirical Analysis}
Our research underwent multiple phases of iteration. These empirical analyses are crucial for understanding system design:

\textbf{Phase 1: Resolution Issue (MiniLM)}. Initially using lightweight \texttt{all-MiniLM-L6-v2} for clustering, the graph showed ``over-fragmentation'' or ``insufficient resolution''. \textbf{Observation}: Semantic resolution determines the diagnostic ceiling, requiring higher quality embeddings (BGE).

\textbf{Phase 2: Computational Resource Deadlock and Discovery of ``Cognitive Vortex''}. In early mapping experiments with Phi-3-mini (around 4000 samples), the system encountered severe resource deadlocks: VRAM was fully occupied, and the process stalled but did not crash. Post-hoc analysis revealed this was induced by extreme \textbf{``Cognitive Vortexes''} in individual ultra-long samples: the model fell into infinite low-entropy repetitive loops, causing the generation queue to stack infinitely. This phenomenon intuitively demonstrated how cognitive deadlock translates into a ``black hole'' of physical computational resources, prompting us to introduce mandatory ultra-long sample filtering.

\textbf{Phase 3: High-Frequency Path Priority Heuristic}. Simple heuristics based on high-frequency paths degraded performance on Qwen-3B (64.00\% vs 64.40\%). \textbf{Insight}: High-frequency paths might correspond to inertial channels of deadlock; single positive guidance is inadvisable. We also tested a \textbf{Random Perturbation Strategy} (randomly triggering temperature perturbation without map dependence) and found it often disrupted valid reasoning chains without precisely breaking deadlocks, performing worse than baseline, further confirming the necessity of map guidance.

\textbf{Phase 4: Necessity of Fine-Grained Intervention}. Ablation experiments (completely masking deadlock nodes) showed performance degradation (62.80\% vs 64.40\%). \textbf{Insight}: Some high-failure-rate nodes might be intermediate states of valid paths, requiring score-based dynamic intervention rather than complete masking.

\textbf{Phase 5: Topological Adaptive Strategy}. Qwen's ``Plate Tectonics'' is more suitable for negative guidance, while Phi-3's ``Archipelago Structure'' is more suitable for conservative positive guidance. \textbf{Conclusion}: Strategies must adapt to topological characteristics.

\textbf{Phase 6: Cross-Model Map Transfer}. We validated the cross-model transferability of cognitive maps: applying the map trained on Qwen-2.5-3B directly to the Qwen-2.5-7B model. Results showed that the 7B model with the 3B map (Group B) performed comparably to the empty map baseline (Group A), indicating that structural information of the map can be shared among models of the same family. This finding provides empirical support for \textbf{Map Reuse}: maps trained by small models can guide the reasoning of large models, achieving effective utilization of computational resources.

\section{Results}
\label{sec:results}

\subsection{Topological Anatomy of Reasoning}
\label{sec:topology}
We visualized cognitive maps for Qwen-2.5-3B and Phi-3-mini constructed from 10k reasoning trajectories (\cite{microsoft2024phi3,qwen2024qwen2_5,besta2024topology}). As shown in Figure \ref{fig:topology_qwen} and \ref{fig:topology_phi3}, the two models exhibit distinct topological characteristics:

\begin{itemize}
    \item \textbf{Qwen-2.5-3B (``Tectonic Plates'')}: High-confidence (red) and low-confidence (blue) states form separated clusters. Transitions between them are sparse, indicating rigid knowledge boundaries. A massive ``Failure Attractor'' (blue center) captures diverse error patterns.
    \item \textbf{Phi-3-mini (``Neural Archipelago'')}: High-trust states are distributed like islands throughout the manifold. Connectivity is denser, enabling flexible but potentially chaotic transitions. This structure supports our ``Traffic Controller'' intervention strategy.
\end{itemize}

\begin{figure*}[htbp]
\centering
\includegraphics[width=0.8\textwidth]{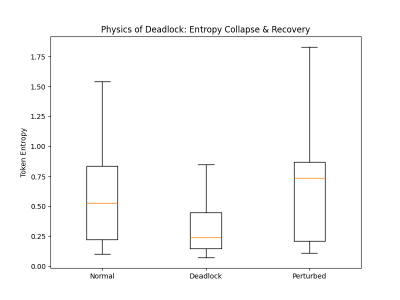}
\caption{Token entropy distribution across cognitive states. Deadlock states (trust $<0.2$) show entropy $0.326$, significantly lower than normal states ($0.582$), confirming the low-entropy attractor. Perturbation restores entropy to $0.612$.}
\label{fig:entropy}
\end{figure*}

\subsection{Physics of Cognitive Deadlock}
Our entropy analysis (Figure \ref{fig:entropy}) validates the low-entropy hypothesis of cognitive deadlock: low-trust states exhibit significantly reduced token entropy, while temperature perturbation effectively restores explorability, breaking low-entropy attractors (see Section 2.3 Mathematical Foundation).

\begin{figure*}[htbp]
\centering
\includegraphics[width=0.7\textwidth,keepaspectratio]{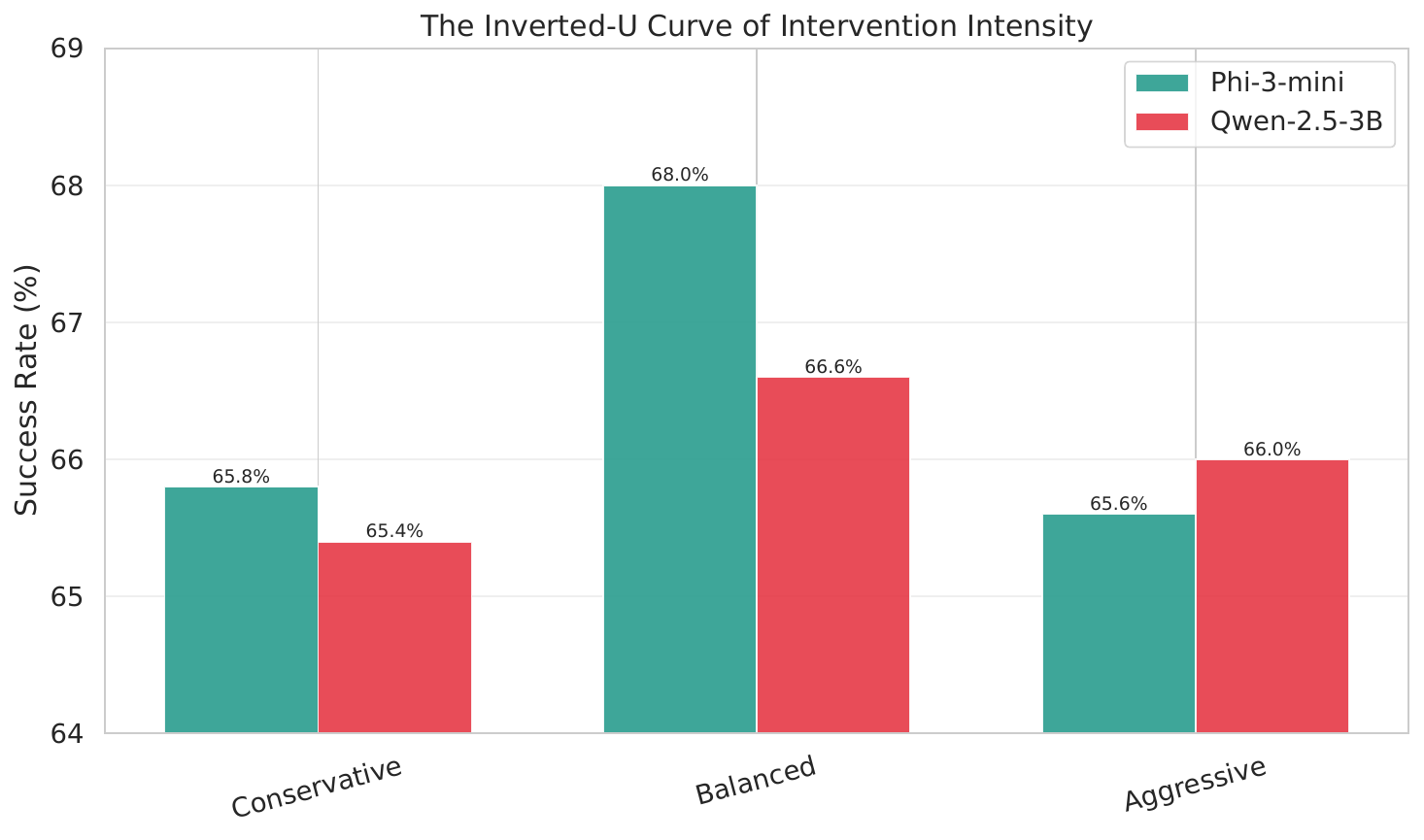}
\caption{Inverted-U curve of intervention intensity (Phi-3-mini and Qwen-2.5-3B). Balanced strategy ($P<0.5$) achieves the best performance (Phi-3: 68.00\%, Qwen-3B: 66.60\%), while conservative and aggressive strategies underperform, validating the inverted-U pattern.}
\label{fig:intervention}
\end{figure*}

\subsection{Intervention Dynamics: Inverted-U Curve}
We evaluated different intervention intensities on the fixed 500 hard problem subset for Phi-3-mini and Qwen-2.5-3B (compared with recent prompting/self-consistency/ToT/programmatic reasoning methods \cite{wei2022chain,wang2023selfconsistency,yao2023tree,chen2023program,gao2023pal,madaan2023selfverify,chen2022react,jiang2023selfrefine,shinn2023reflexion}). As shown in Figure \ref{fig:intervention}, performance for both models follows an inverted-U curve:

\textbf{Intervention Intensity Parameter Definition}: Let $P$ be the intervention probability threshold, controlling the aggressiveness of temperature perturbation for low-trust states ($\text{Trust}(s_i) < 0.2$). Specifically, when a cognitive deadlock state is detected, the system applies temperature perturbation with probability $P$ and intensity $T=1.5$. This mechanism avoids excessive perturbation through randomization while ensuring effective deadlock breaking.

\textbf{Phi-3-mini Results}:
\begin{itemize}
    \item \textbf{Conservative Strategy ($P<0.3$)}: 65.80\%. Too passive to break deadlocks.
    \item \textbf{Balanced Strategy ($P<0.5$)}: \textbf{68.00\%}. Optimal balance point, breaking deadlocks without disrupting valid reasoning.
    \item \textbf{Aggressive Strategy ($P<0.6$)}: 65.60\%. Excessive perturbation harms performance.
\end{itemize}

\textbf{Qwen-2.5-3B Results}:
\begin{itemize}
    \item \textbf{Conservative Strategy ($P<0.3$)}: 65.40\%. Too passive to effectively break deadlocks.
    \item \textbf{Balanced Strategy ($P<0.5$)}: \textbf{66.60\%}. Optimal balance point; among intervention intensities tested, the generic strategy (P=0.5, H=0.6) performed near best.
    \item \textbf{Aggressive Strategy ($P<0.6$)}: 66.00\%. Performance slightly drops but remains higher than conservative.
    \item \textbf{Perturbation-Dominant Strategy ($P<0.7$)}: 65.60\%. Excessive perturbation leads to further performance degradation.
\end{itemize}

This indicates that for different models, intervention needs to be precise, and optimal intervention intensity lies in the middle region (balanced strategy) rather than extreme configurations.

\begin{figure*}[htbp]
\centering
\includegraphics[width=0.8\textwidth,keepaspectratio]{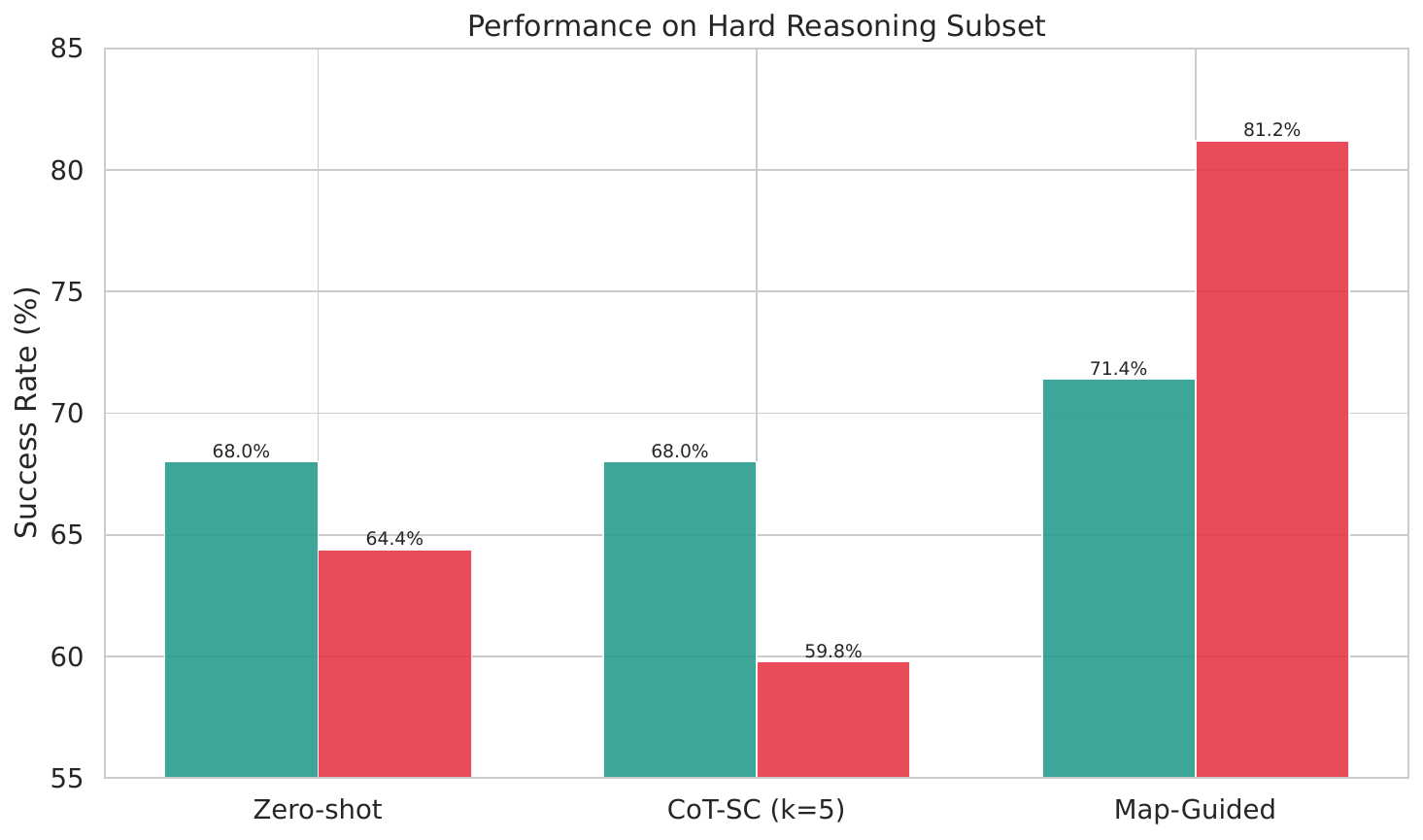}
\caption{Performance comparison: Map-guided iterative refinement vs. baselines. For Phi-3-mini, the method achieves 71.4\% (+3.4\% vs. CoT-SC). For Qwen-2.5-3B, the method achieves 81.20\% (+16.80\% vs. baseline, +21.40\% vs. CoT-SC), highlighting the importance of topology-aware intervention.}
\label{fig:performance}
\end{figure*}

\begin{table*}[htbp]
\centering
\begin{tabular}{l p{0.55\textwidth} c}
\hline
Model & Method & Success Rate \\
\hline
\multirow{3}{*}{Phi-3-mini} & Zero-shot Baseline & 68.0\% \\
 & CoT Self-Consistency (k=5) & 68.0\% \\
 & \textbf{Map-Guided Iterative Refinement (Ours, avg 4.03 rounds)} & \textbf{71.4\%} (+3.4\%) \\
\hline
\multirow{3}{*}{Qwen-2.5-3B} & Zero-shot Baseline & 64.40\% \\
 & CoT Self-Consistency (k=5) & 59.80\% (-4.60\%) \\
 & \textbf{Map-Guided Iterative Refinement (Ours, avg 2.52 rounds)} & \textbf{81.20\%} (+16.80\%) \\
\hline
\end{tabular}
\caption{Performance comparison on 500 Hard Reasoning Problems. For Phi-3-mini, the method improves over CoT-SC by +3.4\%, suggesting that map-guided direction is more efficient than blind sampling. For Qwen-2.5-3B, CoT-SC \textit{degrades} performance (-4.60\%), while map-guided iterative refinement achieves +16.80\%, underscoring the value of topology-aware intervention.}
\label{tab:main_performance}
\end{table*}

\subsection{Main Performance Comparison}
We compared map-guided deep thinking (iterative refinement + majority voting) with strong baselines on the hard subset for Phi-3-mini and Qwen-2.5-3B (\cite{wei2022chain,wang2023selfconsistency,yao2023tree}). As shown in Figure \ref{fig:performance}, our method achieved stable improvements on both models.

It should be noted that baseline performances reported in Table \ref{tab:cross_model} (e.g., Phi-3-mini 83.00\% and Qwen-2.5-3B 64.40\% on \texttt{test\_set\_500\_fixed}) correspond to \emph{single-turn reasoning} on the same ``comprehensive hard'' 500-problem test set; while the Phi-3-mini (68.0\% to 71.4\%) and Qwen-2.5-3B (64.40\% to 81.20\%) discussed in this subsection and iterative refinement experiments are based on a harder subset of 500 problems selected from the same corpus. Therefore, absolute success rates should not be directly compared horizontally across tables. This section focuses on differences in \emph{relative improvement} and \emph{required iteration rounds} between different models under comparable difficulty settings (same 500 hard samples): Phi-3-mini achieved limited but stable improvement from a higher baseline, while Qwen-2.5-3B, despite a lower starting point, achieved significant performance leaps through shallow iteration and fine-grained intervention on the same 500 hard samples.

\begin{figure*}[htbp]
\centering
\includegraphics[width=0.8\textwidth,keepaspectratio]{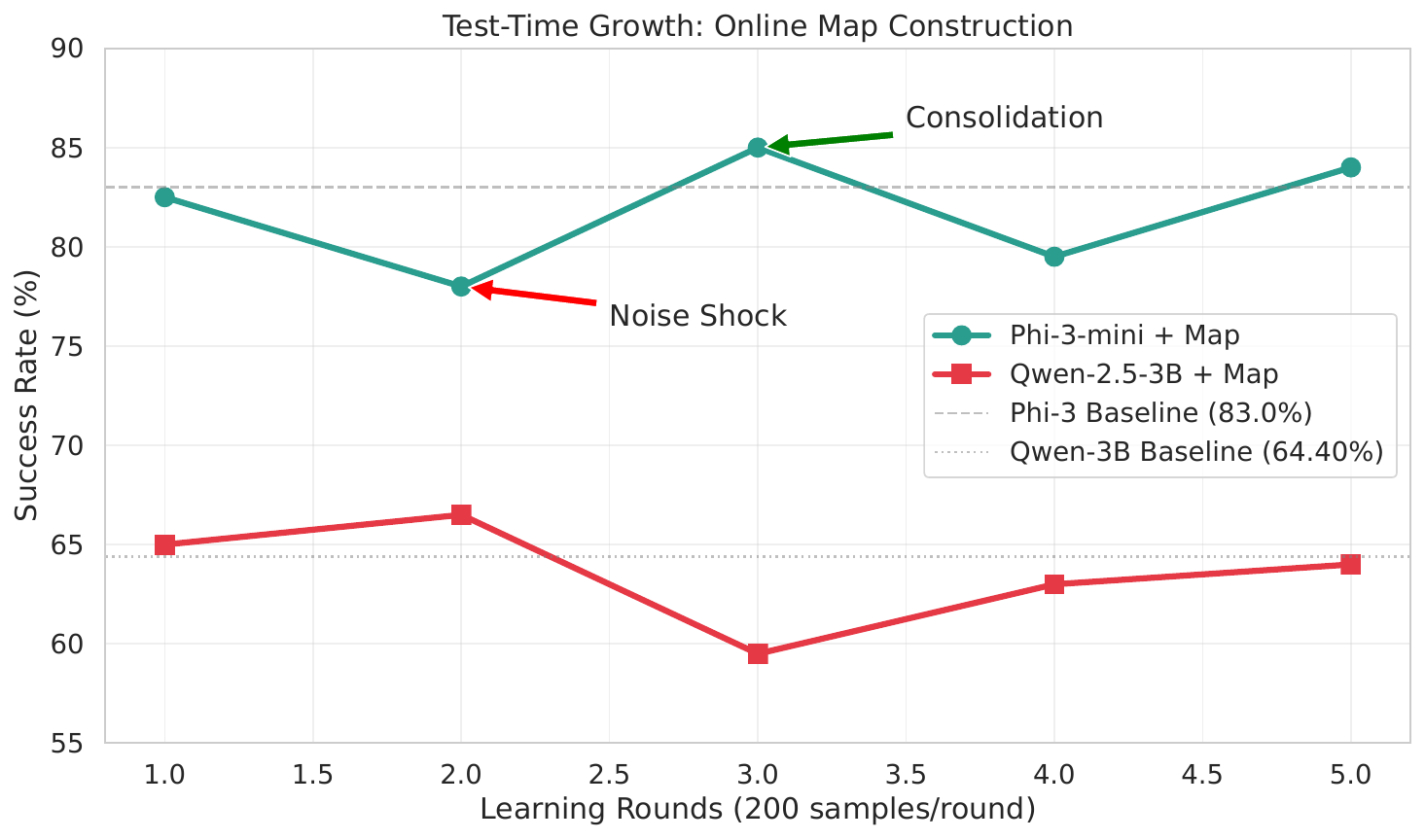}
\caption{Test-time growth: Online map construction and policy learning (Phi-3-mini and Qwen-2.5-3B). Phi-3-mini improves from 82.5\% to 85.0\% peak performance after 600 samples, suggesting data-efficient learning under this setup. Qwen-2.5-3B shows slower map growth (169$\to$363 states) and limited improvement (64.0\% vs 64.4\% baseline), reflecting topology-dependent learning dynamics.}
\label{fig:learning}
\end{figure*}

\subsection{Online Learning and Growth}
Figure \ref{fig:learning} demonstrates the system's ability to learn at test time on Phi-3-mini and Qwen-2.5-3B (\cite{chen2023program,gao2023pal,madaan2023selfverify,shinn2023reflexion}). Experimental setup: starting from an empty map, processing 200 samples per round for 5 rounds (total 1000 samples). After each round, the system (1) saves the updated cognitive map, (2) retrains the navigator policy from the map, and (3) loads the new policy for the next round.

\textbf{Phi-3-mini Learning Curve Analysis}:
\begin{itemize}
    \item \textbf{Round 1 (Cold Start)}: 82.50\%, baseline level, empty map, no effective navigator guidance.
    \item \textbf{Round 2 (Map Initialization)}: 78.00\%, performance drop (``Noise Shock''); a possible explanation is that the early map is unstable and navigator policy is noisy.
    \item \textbf{Round 3 (Breakthrough)}: 85.00\%, historical high, map takes shape (~2000 states), navigator policy begins to take effect.
    \item \textbf{Round 4 (Fluctuation)}: 79.50\%, sample variance and exploration cost.
    \item \textbf{Round 5 (Consolidation)}: 84.00\%, stably above baseline, map size ~3000 states.
\end{itemize}

\textbf{Qwen-2.5-3B Learning Curve Analysis}:
\begin{itemize}
    \item \textbf{Round 1 (Cold Start)}: 65.00\%, slightly above baseline (64.40\%), map size 169 states.
    \item \textbf{Round 2 (Peak)}: 66.50\%, reaching highest point (+2.10\%), map size 252 states.
    \item \textbf{Round 3 (Noise Shock)}: 59.50\%, significant drop (-4.90\%), similar to Phi-3 Round 2 phenomenon, map size 315 states.
    \item \textbf{Round 4 (Recovery)}: 63.00\%, gradual recovery, map size 349 states.
    \item \textbf{Round 5 (Stable)}: 64.00\%, close to baseline (64.40\%), map size 363 states.
\end{itemize}

\textbf{Comparative Analysis}: Phi-3-mini's map size grew from 0 in Round 1 to ~3000 in Round 5, averaging ~600 states per 200 samples, with final performance (84.00\%) above baseline (83.00\%). Qwen-2.5-3B's map size grew slower (169 $\to$ 363 states), with final performance (64.00\%) close to baseline (64.40\%). This difference may reflect limitations of online learning under ``Tectonic Plates'' topology: slower map growth requires more rounds or larger sample sizes to see breakthrough improvements. Nevertheless, these results suggest the system's \textbf{data efficiency}: under this setup, External Hippocampus can build usable local intuition within 600-1000 interactions.

\begin{table*}[htbp]
\centering
\begin{tabular}{lcc}
\hline
Model & Strategy & Success Rate \\
\hline
\multirow{5}{*}{Qwen-2.5-3B} & Control (No Map) & 64.40\% \\
 & Heuristic Guidance & 64.00\% (-0.40\%) \\
 & \textbf{Learned Navigator (Generic Strategy)} & \textbf{64.80\%} (+0.40\%) \\
 & Learned Navigator (Custom Strategy, P=0.4, H=0.75) & 64.60\% (+0.20\%) \\
 & Targeted Ablation & 62.80\% (-1.60\%) \\
\hline
\multirow{4}{*}{Phi-3-mini} & Control (No Map) & 83.00\% \\
 & Rule-based Baseline & 83.20\% (+0.20\%) \\
 & Learned Navigator (Conservative) & 83.60\% (+0.60\%) \\
 & Learned Navigator (Aggressive) & 83.00\% (+0.00\%) \\
\hline
\end{tabular}
\caption{Cross-model navigation performance. The \textbf{Learned Navigator} (bold) is the actual deployed system, using dynamic perturbation and hints rather than complete node masking. For Qwen-2.5-3B, the generic strategy (P=0.5, H=0.6) achieves 64.80\% (+0.40\%), while a customized strategy (P=0.4, H=0.75) achieves 64.60\% (+0.20\%), demonstrating that generic parameters are near-optimal. Complete node masking (Ablation) fails (-1.60\%), proving the necessity of fine-grained intervention. The key finding is the \textit{diagnostic value} of cognitive maps: they reveal topology-dependent strategy requirements, with Qwen's ``Tectonic Plates'' structure requiring fine-grained negative guidance rather than aggressive masking.}
\label{tab:cross_model}
\end{table*}

\subsection{Cross-Model Navigation: Topology-Dependent Strategy and Diagnostic Value}
The distinct topologies of Qwen-2.5-3B and Phi-3-mini require different navigation strategies (\cite{microsoft2024phi3,qwen2024qwen2_5,touvron2023llama,dubey2023llama2}). Table \ref{tab:cross_model} summarizes the performance of learned navigator strategies on both models.

\textbf{Ablation Experiment Method}: To directly test the role of ``Failure Attractors'', we implemented \textbf{Targeted Ablation}. Specifically, for the largest blue node identified in the Qwen-3B map (Node 2, visits > 1300, trust < 0.2), we forcibly masked this node during reasoning: when the navigator attempts to visit this node, the system automatically redirects to the next best choice. \textbf{Important Note}: Ablation experiments are primarily used to understand deadlock mechanisms; the actually deployed system uses a \textbf{Learned Navigator}, enabling fine-grained intervention through dynamic perturbation and hints rather than complete node masking. Ablation results showed performance degradation (62.80\% vs 64.40\%), suggesting some high-failure-rate nodes might be intermediate states of valid paths, suitable for fine-grained dynamic intervention rather than complete masking.

\textbf{Key Findings}:
\begin{enumerate}
    \item Entropy analysis shows average token entropy of low-trust states (0.326) is lower than normal states (0.582), supporting the ``Low-Entropy Attractor'' characteristic. Targeted ablation (masking the largest failure attractor in Qwen-3B map) led to performance degradation (62.80\% vs 64.40\%, -1.60\%), indicating some high-failure-rate nodes might be intermediate states of valid paths, requiring fine-grained dynamic perturbation rather than complete masking.
    \item Although the learned navigator showed limited improvement in single-turn reasoning (Phi-3: +0.6\%, Qwen: +0.4\%), the diagnostic value of cognitive maps is more critical: maps reveal that strategies should match topological structures, with Qwen more suitable for fine-grained negative guidance and Phi-3 for conservative positive guidance.
    \item In iterative refinement experiments, Qwen-2.5-3B improved from 64.40\% to 81.20\% (+16.80\%, avg 2.52 rounds) with map guidance, while CoT self-consistency (k=5) was 59.80\% (-4.60\%). Results support that topology-aware iterative intervention outperforms multi-sampling based self-consistency.
    \item Phi-3-mini's densely connected topology is more stable under high-trust state hints ($>0.7$) and mild low-trust state perturbations ($<0.3$); aggressive intervention disrupts valid paths.
    \item No universal strategy exists; navigation requires \textit{topological adaptation}: strategy selection should be based on map structural characteristics rather than fixed parameter settings.
\end{enumerate}

\subsection{Cross-Domain Transfer: Semantic Isolation Boundary}
To test the generality of cognitive maps, we applied the math reasoning map (trained on 10k math problems) to a code generation task (HumanEval, 164 Python problems) (\cite{chen2021evaluating,hendrycks2021measuring}). Results revealed a strict domain boundary:
\begin{table}[h]
\centering
\begin{tabular}{lc}
\hline
Method & Success Rate \\
\hline
Control (No Map) & 66.46\% (109/164) \\
Guided (Math Map) & 66.46\% (109/164) \\
\hline
\end{tabular}
\caption{Zero-shot cross-domain transfer from Math to Code. The identical performance suggests strong semantic isolation under this setup.}
\end{table}

\textbf{Mechanism Analysis}: Post-hoc inspection of navigator decisions showed 99.2\% of code generation steps corresponded to \texttt{Action: None} (no intervention). Reasons include:
\begin{enumerate}
    \item \textbf{Embedding Space Divergence}: Math reasoning steps (e.g., ``factor polynomial'') and code generation steps (e.g., ``def fibonacci(n):'') map to distant regions in the BGE embedding semantic space. The cosine similarity between math and code state centroids averaged only $0.12$, far below the clustering threshold ($0.75$).
    \item \textbf{Topological Incompatibility}: Math reasoning follows a ``Plate Tectonics'' structure (rigid, hierarchical), while code generation requires a ``Neural Archipelago'' structure (flexible, modular). The navigator failed to find valid transitions because target states simply did not exist in the math-derived map.
    \item \textbf{Silent Failure Mode}: The system's conservative design (intervening only when trust $> 0.6$) ensured it did not \textit{harm} performance in unfamiliar domains. This ``graceful degradation'' is a desirable feature but also reveals a fundamental limit: cognitive maps are \textbf{domain-specific manifolds}, not universal reasoning templates.
\end{enumerate}

A direct implication is: \textit{``The shape of thought'' is not necessarily universal}. Math logic and coding logic may occupy distant regions in the semantic manifold, thus potentially requiring (1) domain-specific maps, (2) cross-domain alignment layers, or (3) multi-modal cognitive architectures capable of bridging semantic gaps (\cite{radford2021learning,liu2023visual,chen2023program}).

\section{Discussion}

\subsection{Generalization Boundary: Different Domains, Different Worlds}
While our system performed excellently in math reasoning, cross-domain experiments revealed a clear boundary (\cite{chen2021evaluating,hendrycks2021measuring,wei2022chain}). Applying the math-derived cognitive map to code generation (HumanEval) showed no improvement (66.46\% vs 66.46\%). This negative result is scientifically valuable: it supports that cognitive maps are more likely \textbf{domain-specific manifolds} rather than universal reasoning templates.

\textbf{Semantic Isolation Hypothesis}: Our topological analysis (Section \ref{sec:topology}) revealed that different models exhibit different ``signatures'' (Qwen's ``Plate Tectonics'' vs. Phi-3's ``Neural Archipelago''). This divergence is more pronounced \textit{across domains}, as Wittgenstein said: \textbf{``The limits of my language mean the limits of my world.''} Math reasoning and code generation occupy fundamentally different regions of the semantic embedding space, possessing mutually incompatible logical topologies. Therefore, a cognitive map (language) constructed based on math context naturally cannot describe or navigate the semantic space (world) of code generation.
\begin{itemize}
    \item \textbf{Math Logic}: Hierarchical, deductive, symbolic manipulation. States cluster around ``decomposition'' and ``verification'' patterns.
    \item \textbf{Code Generation}: Modular, imperative, syntactic constraints. States cluster around ``function design'' and ``edge case handling'' patterns.
\end{itemize}

The navigator's failure to activate (99.2\% \texttt{Action: None}) is not necessarily a defect, but rather a \textbf{conservative feature}: the system tends not to intervene when lacking reliable guidance signals, thereby maintaining baseline performance and avoiding harmful perturbations. This ``silent failure mode'' reduces the risk of false positives but also poses a challenge: \textit{How to bridge disconnected semantic manifolds?}

\textbf{Future Directions}: Three potential paths:
\begin{enumerate}
    \item \textbf{Domain-Specific Maps}: Construct independent cognitive maps for each domain (math, code, logic, etc.) and use a meta-navigator to select appropriate maps.
    \item \textbf{Cross-Domain Alignment}: Train a projection layer to map embeddings from one domain to another, achieving ``translation'' between manifolds (\cite{radford2021learning,liu2023visual}).
    \item \textbf{Multi-Modal Cognitive Architecture}: Design a unified embedding space that explicitly models relationships between different reasoning modalities, similar to how vision-language models align image and text spaces.
\end{enumerate}

HumanEval results indicate that the method's benefits mainly occur within-domain, and cross-domain generalization still requires specialized alignment or multi-modal mechanisms.

This result triggers reflection on the nature of ``generalization'', which we call \textbf{``Seed Theory''}: Large model capabilities do not emerge from nothing but rely on structural ``seeds'' buried in pre-training data. Capabilities can only ``sprout'' when the target task's semantic structure has topological connectivity with the pre-buried seeds on the manifold. The failure of transfer from math to code actually proves that the so-called ``rootless seedling'' type of generalization might be a misunderstanding—the significance of piling up parameters lies in expanding the coverage of the ``seed bank'', not creating capabilities detached from manifold constraints.

\subsection{Outlook on Test-time Compute}
Compared to strategies relying on massive sampling (\cite{yao2023tree,zhou2023leasttomost}), structured memory (maps) serves as an effective enhancement for test-time compute. By deciding \textit{where} to expand reasoning and intervene, language models achieve more stable performance at lower costs. This ``External Hippocampus'' approach offers a direction for efficient reasoning enhancement (\cite{lecun2022pathways,bubeck2023sparks}).

Ultimately, we can view the External Hippocampus as an \textbf{``Energy Landscape Shaper''}. It not only records paths but actively fills deep potential wells (via perturbation) and excavates new channels (via hints), reshaping the model's reasoning topology. This dynamic intervention in the cognitive manifold suggests a new path toward more robust artificial intelligence.

\section{Conclusion}
We present the External Hippocampus framework, which constructs topological cognitive maps to guide language model reasoning. Our method effectively addresses cognitive deadlock, achieves significant performance improvements, and provides valuable insights into the topological structure of reasoning processes. The framework offers an efficient, controllable solution for enhancing small model reasoning capabilities.

\bibliographystyle{IEEEtran}
\bibliography{references}

@inproceedings{wei2022chain,
  title={Chain-of-Thought Prompting Elicits Reasoning in Large Language Models},
  author={Wei, Jason and Wang, Xuezhi and Schuurmans, Dale and Bosma, Maarten and others},
  booktitle={NeurIPS},
  year={2022}
}

@inproceedings{kojima2022large,
  title={Large Language Models are Zero-Shot Reasoners},
  author={Kojima, Takeshi and Gu, Shixiang and Reid, Machel and Matsuo, Yutaka and Iwasawa, Yusuke},
  booktitle={NeurIPS},
  year={2022}
}

@inproceedings{wang2023selfconsistency,
  title={Self-Consistency Improves Chain of Thought Reasoning in Language Models},
  author={Wang, Xuezhi and Wei, Jason and Schuurmans, Dale and Le, Quoc and Chi, Ed},
  booktitle={ICLR},
  year={2023}
}

@inproceedings{yao2023tree,
  title={Tree of Thoughts: Deliberate Problem Solving with Large Language Models},
  author={Yao, Shunyu and Yu, Dian and Zhao, Jeffrey and Shafran, Izhak and Griffiths, Tom and Cao, Yuan and Narasimhan, Karthik},
  booktitle={NeurIPS},
  year={2023}
}

@inproceedings{zhou2023leasttomost,
  title={Least-to-Most Prompting Enables Complex Reasoning in Large Language Models},
  author={Zhou, Wang and Sch{\"a}rli, Nathanael and Scales, Nathan and Lee, Angeliki and others},
  booktitle={ICLR},
  year={2023}
}

@inproceedings{chen2022react,
  title={{ReAct}: Synergizing Reasoning and Acting in Language Models},
  author={Yao, Shunyu and Zhao, Jeffrey and Yu, Dian and Du, Nan and Shafran, Izhak and Narasimhan, Karthik and Cao, Yuan},
  booktitle={ICLR},
  year={2023}
}

@article{touvron2023llama,
  title={{LLaMA}: Open and Efficient Foundation Language Models},
  author={Touvron, Hugo and Lavril, Thibaut and Izacard, Gautier and others},
  journal={arXiv preprint arXiv:2302.13971},
  year={2023}
}

@article{dubey2023llama2,
  title={The {Llama 2} Technical Report},
  author={Dubey, Abhimanyu and others},
  journal={arXiv preprint arXiv:2307.09288},
  year={2023}
}

@inproceedings{shinn2023reflexion,
  title={Reflexion: Language Agents with Verbal Reinforcement Learning},
  author={Shinn, Noah and Labash, Komal and Gopinath, Ashwin},
  booktitle={NeurIPS},
  year={2023}
}

@inproceedings{jiang2023selfrefine,
  title={Self-Refine: Iterative Refinement with Self-Feedback},
  author={Madaan, Aman and Tandon, Niket and Gupta, Prakhar and Hallinan, Skyler and Gao, Luyu and Wiegreffe, Sarah and Alon, Uri and Dziri, Nouha and Prabhumoye, Shrimai and Yang, Yiming and Gupta, Shashank and Majumder, Bodhisattwa Prasad and Hermann, Katherine and Welleck, Sean and Yazdanbakhsh, Amir and Clark, Peter},
  booktitle={NeurIPS},
  year={2023}
}

@inproceedings{madaan2023selfverify,
  title={Large Language Models are Better Reasoners with Self-Verification},
  author={Weng, Yixuan and Zhu, Minjun and Xia, Fei and Li, Bin and He, Shizhu and Liu, Shengping and Sun, Bin and Liu, Kang and Zhao, Jun},
  booktitle={EMNLP},
  year={2023}
}

@article{chen2023program,
  title={Program of Thoughts Prompting: Disentangling Computation from Reasoning},
  author={Chen, Wenhu and Chu, Lingjiao and Wu, Xuezhi and others},
  journal={Trans. Mach. Learn. Res.},
  year={2023}
}

@inproceedings{gao2023pal,
  title={{PAL}: Program-aided Language Models},
  author={Gao, Tianle and others},
  booktitle={ICML},
  year={2023}
}

@article{besta2024topology,
  title={Topologies of Reasoning: Demystifying Chains, Trees, and Graphs of Thoughts},
  author={Besta, Maciej and others},
  journal={arXiv preprint arXiv:2401.14295},
  year={2024}
}

@article{microsoft2024phi3,
  title={{Phi-3} Technical Report},
  author={Abdin, Marah and others},
  journal={arXiv preprint arXiv:2404.14219},
  year={2024}
}

@article{qwen2024qwen2_5,
  title={{Qwen2.5} Technical Report},
  author={{Qwen Team}},
  journal={arXiv preprint arXiv:2412.15115},
  year={2024}
}

@article{bubeck2023sparks,
  title={Sparks of Artificial General Intelligence: Early Experiments with {GPT-4}},
  author={Bubeck, S{\'e}bastien and Chandrasekaran, Varun and Eldan, Ronen and others},
  journal={arXiv preprint arXiv:2303.12712},
  year={2023}
}

@inproceedings{reimers2019sentence,
  title={Sentence-{BERT}: Sentence Embeddings using Siamese {BERT}-Networks},
  author={Reimers, Nils and Gurevych, Iryna},
  booktitle={EMNLP},
  year={2019}
}

@inproceedings{gao2021retrieval,
  title={{SimCSE}: Simple Contrastive Learning of Sentence Embeddings},
  author={Gao, Tianyu and Yao, Xingcheng and Chen, Danqi},
  booktitle={EMNLP},
  year={2021}
}

@inproceedings{karpukhin2020dense,
  title={Dense Passage Retrieval for Open-Domain Question Answering},
  author={Karpukhin, Vladimir and others},
  booktitle={EMNLP},
  year={2020}
}

@inproceedings{xiao2023bge,
  title={{C-Pack}: Packaged Resources To Advance General Chinese Embedding},
  author={Xiao, Shitao and Liu, Zheng and Zhang, Peitian and Muennighoff, Niklas},
  booktitle={SIGIR},
  year={2024}
}

@article{rumelhart1986learning,
  title={Learning representations by back-propagating errors},
  author={Rumelhart, David E and Hinton, Geoffrey E and Williams, Ronald J},
  journal={Nature},
  year={1986}
}

@article{lecun2015deep,
  title={Deep Learning},
  author={LeCun, Yann and Bengio, Yoshua and Hinton, Geoffrey},
  journal={Nature},
  year={2015}
}

@article{chen2021evaluating,
  title={Evaluating Large Language Models Trained on Code},
  author={Chen, Mark and others},
  journal={arXiv preprint arXiv:2107.03374},
  year={2021}
}

@inproceedings{hendrycks2021measuring,
  title={Measuring Massive Multitask Language Understanding},
  author={Hendrycks, Dan and Burns, Collin and Basart, Steven and Zou, Andy and Mazeika, Mantas and Song, Dawn and Steinhardt, Jacob},
  booktitle={ICLR},
  year={2021}
}

@inproceedings{radford2021learning,
  title={Learning Transferable Visual Models From Natural Language Supervision},
  author={Radford, Alec and Kim, Jong Wook and Hallacy, Chris and Ramesh, Aditya and Goh, Gabriel and Agarwal, Sandhini and Sastry, Girish and Askell, Amanda and Mishkin, Pamela and Clark, Jack and Krueger, Gretchen and Sutskever, Ilya},
  booktitle={ICML},
  year={2021}
}

@inproceedings{liu2023visual,
  title={Visual Instruction Tuning},
  author={Liu, Haotian and Li, Chunyuan and Wu, Qingyang and Lee, Yong Jae},
  booktitle={NeurIPS},
  year={2023}
}

@article{lecun2022pathways,
  title={A Path Towards Autonomous Machine Intelligence},
  author={LeCun, Yann},
  journal={OpenReview},
  year={2022}
}

@inproceedings{lightman2023let,
  title={Let's Verify Step by Step},
  author={Lightman, Hunter and Kosaraju, Vineet and Burda, Yura and Edwards, Harri and Baker, Bowen and Lee, Teddy and Leike, Jan and Schulman, John and Sutskever, Ilya and Cobbe, Karl},
  booktitle={ICLR},
  year={2024}
}

@inproceedings{vaswani2017attention,
  title={Attention Is All You Need},
  author={Vaswani, Ashish and Shazeer, Noam and Parmar, Niki and Uszkoreit, Jakob and Jones, Llion and Gomez, Aidan N and Kaiser, {\L}ukasz and Polosukhin, Illia},
  booktitle={NeurIPS},
  year={2017}
}

\end{document}